\documentclass[10pt,twocolumn,letterpaper]{article}

\usepackage[pagenumbers]{cvpr} 

\usepackage{graphicx}

\usepackage{amsmath}
\usepackage{amssymb}
\usepackage{booktabs}

\usepackage{bbm}

\usepackage{multirow}

\usepackage{subcaption}
\usepackage{algorithm}
\usepackage{algpseudocode}
\usepackage{amsmath}

\usepackage{xcolor,colortbl}
\definecolor{Gray}{gray}{0.94}

\usepackage[pagebackref,breaklinks,colorlinks]{hyperref}

\usepackage[capitalize]{cleveref}
\crefname{section}{Sec.}{Secs.}
\Crefname{section}{Section}{Sections}
\Crefname{table}{Table}{Tables}
\crefname{table}{Tab.}{Tabs.}

\def\cvprPaperID{3913} 
\def\confName{CVPR}
\def\confYear{2022}

\makeatletter
\def\thanks#1{\protected@xdef\@thanks{\@thanks
        \protect\footnotetext{#1}}}
\makeatother
\begin{document}


\title{Dual-Curriculum Teacher \\for Domain-Inconsistent Object Detection in Autonomous Driving}
\author{Longhui Yu$^{1}$ \quad Yifan Zhang$^{2}$ \quad Lanqing Hong$^{3}$\textsuperscript{$\dagger$}$\thanks{\textsuperscript{$\dagger$}~Corresponding author: Dr. Lanqing Hong.}$ \quad Fei Chen$^{3}$ \quad Zhenguo Li$^{3}$\\
{\normalsize $^{1}$Peking University \quad $^{2}$National University of Singapore \quad $^{3}$Huawei Noah’s Ark Lab}\\
{\tt\small yulonghui@stu.pku.edu.cn, yifan.zhang@u.nus.edu,}\\ {\tt\small \{honglanqing, chen.f, li.zhenguo\}@huawei.com}
}
\maketitle


\begin{abstract}
Object detection for autonomous vehicles has received increasing attention in recent years, where labeled data are often expensive while unlabeled data can be collected readily, calling for research on semi-supervised learning for this area. Existing semi-supervised object detection (SSOD) methods usually assume that the labeled and unlabeled data come from the same data distribution. In autonomous driving, however, data are usually collected from different scenarios, such as different weather conditions or different times in a day. Motivated by this, we study a novel but challenging domain-inconsistent SSOD problem.
It involves two kinds of distribution shifts among different domains, including (1) data distribution discrepancy, and (2) class distribution shifts, making existing SSOD methods suffer from inaccurate pseudo-labels and hurting model performance.
To address this problem, we propose a novel method, namely Dual-Curriculum Teacher (DucTeacher). Specifically, DucTeacher consists of two curriculums, \emph{i.e.,} (1) domain evolving curriculum seeks to learn from the data progressively to handle data distribution discrepancy by estimating the similarity between domains, and (2) distribution matching curriculum seeks to estimate the class distribution for each unlabeled domain to handle class distribution shifts. In this way, DucTeacher can calibrate biased pseudo-labels and handle the domain-inconsistent SSOD problem effectively. DucTeacher shows its advantages on SODA10M, the largest public semi-supervised autonomous driving dataset, and COCO, a widely used SSOD benchmark. Experiments show that DucTeacher achieves new state-of-the-art performance on SODA10M  with 2.2 mAP improvement and on COCO with 0.8 mAP improvement.

\end{abstract}

\vspace{-0.40cm}
\section{Introduction}
\label{sec:intro}

\begin{figure}[t]
\centering
\subcaptionbox{Classical Semi-Supervised Object Detection. \label{subfig:classical}}
    {%
        \includegraphics[width=3.1in]{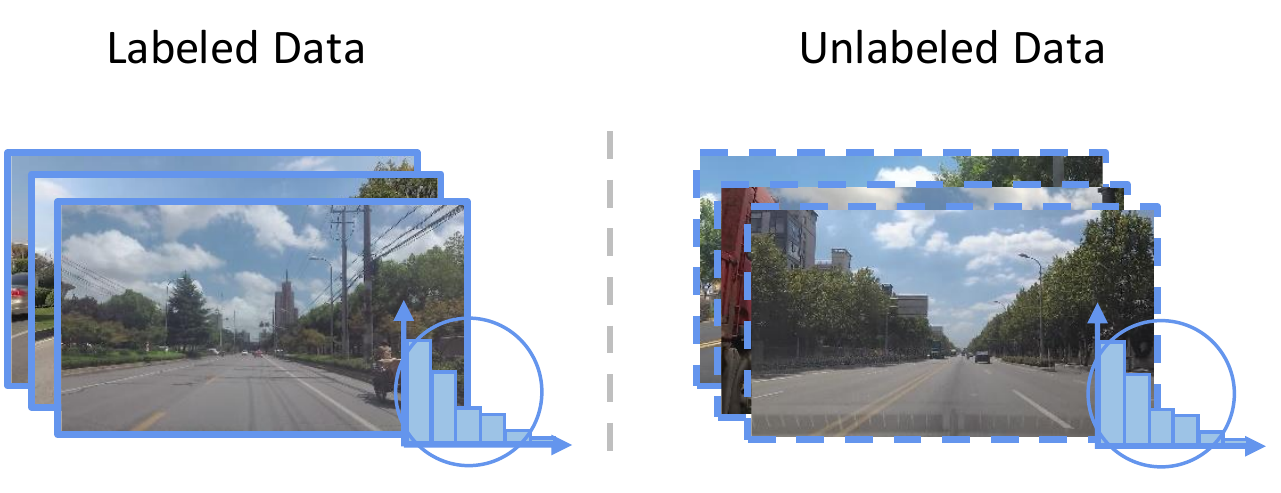}}
\subcaptionbox{Domain-Inconsistent Semi-Supervised Object Detection.\label{subfig:inconsistent}}
    {%
        \includegraphics[width=3.1in]{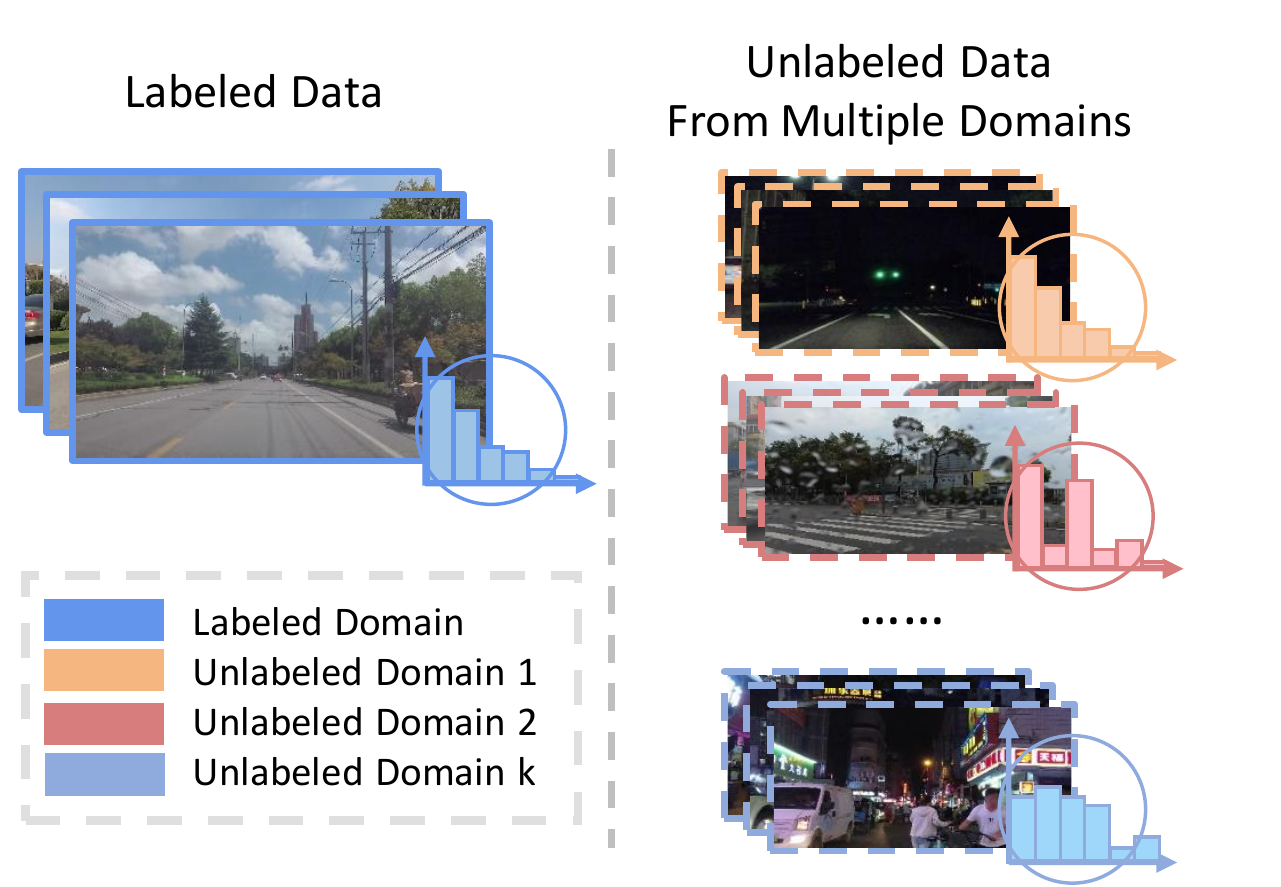}}
\caption{Illustration of domain-inconsistent semi-supervised object detection (SSOD). (a) Existing SSOD often considers labeled data and unlabeled data from the same data distribution. (b) Domain-inconsistent SSOD aims to tackle the problem with both data distribution shifts and class distribution shifts between labeled and unlabeled data. \label{Domain-Inconsistent}}
\vspace{-0.16cm}
\end{figure}
\vspace{-0.16cm}

Autonomous driving~\cite{janai2020computer, grigorescu2020survey, pendleton2017perception} has received considerable attention in recent years because of its potential to ease congestion, reduce emissions, and even save lives.
However, the timeline for the real-world application of autonomous driving is still uncertain due to the unsatisfactory model performance.
One main reason for the limited model performance is the limited size of labeled data, as collecting a large amount of annotated data is usually expensive, especially tasks such as object detection and segmentation. 
Semi-supervised learning (SSL) \cite{sohn2020fixmatch, berthelot2019mixmatch, berthelot2019remixmatch, cascante2020curriculum, he2021re, zhang2021flexmatch}, a paradigm to use abundant unlabeled data to improve the model performance with limited annotated data, is a promising technology for autonomous driving.

\begin{figure*}[t]
\centering
\includegraphics[scale=0.49]{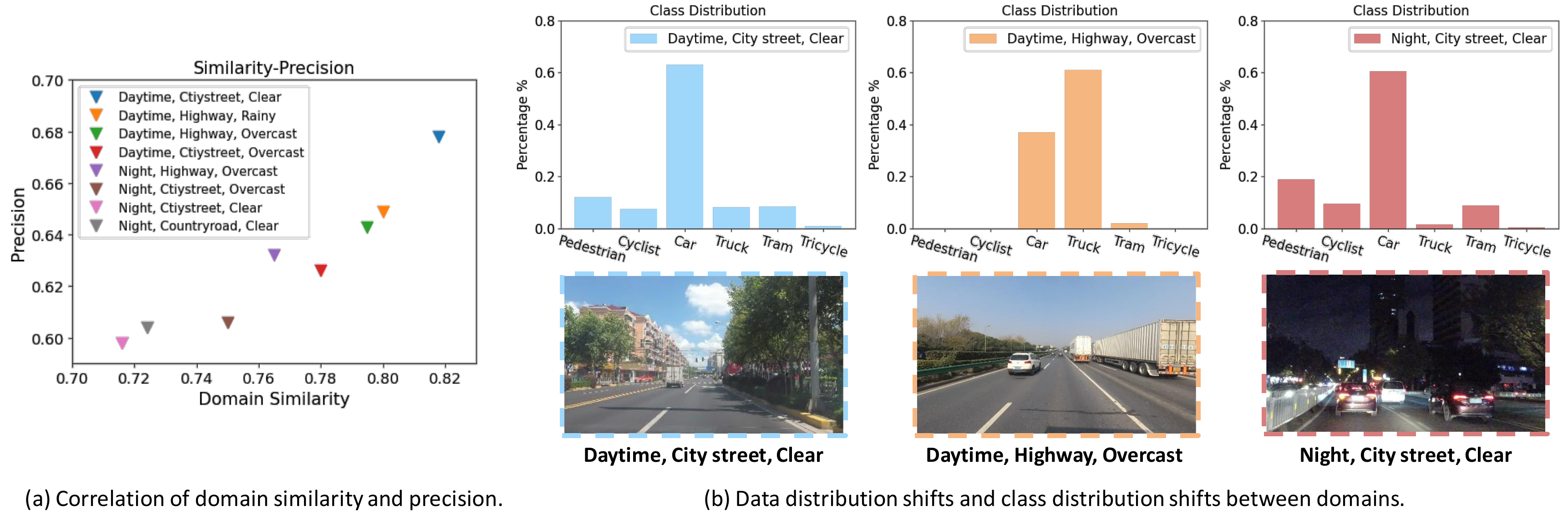}
\caption{(a) Correlation between domain similarity and precision of pseudo-labels. The precision of pseudo-labels produced by the model with Unbiased Teacher \cite{liu2021unbiased} is positively associated with the proposed domain similarity provided by DucTeacher. (b) Data distribution shifts and class distribution shifts between domains. The first row shows the different class distributions in different domains. The second row shows data from different domains that exist pixel-level distribution shifts.} 
\label{intuition}
\end{figure*}

Object detection is one of the most important tasks in autonomous driving, which provides positioning and classification for crucial targets (e.g., pedestrians) for subsequent route planning~\cite{jeong2019consistency, sohn2020simple, zhou2021instant, liu2021unbiased}.
Existing semi-supervised object detection (SSOD) mainly builds upon well-collected datasets, such as ImageNet ~\cite{deng2009imagenet} and COCO~\cite{lin2014microsoft}, assuming that labeled data and unlabeled data are independent and identically distributed (IID).
In autonomous driving, however, data are usually collected from various scenarios, such as different weather conditions or different times in a day, resulting in the data with different distributions.
Motivated by the largest publicly available semi-supervised autonomous driving dataset, SODA10M~\cite{han2021soda10m}, we aim for a novel setting of SSOD, where there are (1) multiple domains in the data and (2) class distribution shifts among the domains.
We name it as \textbf{domain-inconsistent SSOD}. See Fig.~\ref{Domain-Inconsistent} and Fig.~\ref{intuition}(b) as illustrations.

The above domain-inconsistent setting introduces two challenges for SSOD. One is how to tackle the data distribution shift between labeled and unlabeled data. This problem has been investigated in domain adaption~\cite{tzeng2017adversarial,pan2010domain,ganin2015unsupervised,long2015learning,zhang2020collaborative} but is usually neglected in semi-supervised learning~\cite{tarvainen2017mean,xie2019unsupervised,sohn2020fixmatch,berthelot2019mixmatch,berthelot2019remixmatch,cascante2020curriculum,he2021re,zhang2021flexmatch}, which would result in noisy pseudo-labels and hurt the model performance. 
The second challenge is how to track the class distribution shifts in multiple domains.
In SSL, the class distribution of the labeled data is usually adopted as a prior to calibrating the pseudo-label distribution of the unlabeled data~\cite{berthelot2019remixmatch, he2021re}. 
However, with class distribution shifts, existing distribution correction methods~\cite{berthelot2019remixmatch, he2021re}, taking a false class distribution as a reference, would bias to the head classes in the labeled data and result in biased pseudo-labels.
See Sec.~\ref{sec:setting} for more discussions on the challenges.

In this work, we propose Dual-Curriculum Teacher (DucTeacher) for the challenging domain-inconsistent SSOD with data distribution shift and class distribution shift. 
Two curriculum strategies, i.e., distribution matching curriculum (DMC) and domain evolving curriculum (DEC), are proposed to obtain appropriate pseudo-labels for semi-supervised learning.
Specifically, DEC is proposed to learn unlabeled data from different domains progressively to alleviate noisy pseudo-labels caused by the data distribution shift.
It proposes a difficulty metric to measure the domain similarity and introduce unlabeled data from different domains according to the domain similarity.
Moreover, DMC is proposed to adjust the class-specific and domain-adaptive thresholds for pseudo-labeling.
It dynamically adjusts the thresholds for each class, to avoid introducing too many head class pseudo-labels or too little tail class pseudo-label by maintaining an unbiased pseudo-label distribution. Extensive experiments on SODA10M~\cite{han2021soda10m} and COCO~\cite{lin2014microsoft} show the superiority of the proposed DucTeacher.
Overall, the main contributions of this work are three-fold:
\begin{itemize}
\item We target the novel but challenging domain-inconsistent SSOD setting for practical autonomous driving, where the labeled data and unlabeled data come from different domains. Meanwhile, both the data distribution shifts and class distribution shifts happen on the data. 
\item We propose DucTeacher with two curriculum strategies, DEC and DMC, to provide accurate and unbiased pseudo-labels and improve the performance for semi-supervised object detection.
\item In DucTeacher, we develop a novel class distribution estimation method to resist the class distribution shift on the unlabeled data, and a difficulty metric to estimate the domain similarity of unlabeled data from different domains.
\end{itemize}

\section{Related Work}

\noindent \textbf{Curriculum Learning.}
Curriculum learning \cite{bengio2009curriculum} is a learning strategy inspired by the human learning process, which aims to learn training samples from easy to hard. Prior works \cite{wu2020curricula,jiang2015self,cascante2020curriculum,zhang2017curriculum,choi2019pseudo} have shown that curriculum learning can optimize the learning process and improve performance, especially when the label is noisy or the training cost is limited. 
However, how to design a curriculum learning strategy in Autonomous Driving for better training effective is under development

\noindent \textbf{Semi-supervised Object Detection.} Considering the high cost of collecting annotated data, many semi-supervised methods have been proposed for the object detection task~\cite{sohn2020simple,liu2021unbiased}.
For example, STAC~\cite{sohn2020simple} uses a small set of labeled data to pre-train a detector and then uses the detector to generate pseudo-labels for unlabeled data. 
Unbiased Teacher~\cite{liu2021unbiased} improves the pseudo-label generation by using a teacher-student mutual learning framework and alleviates the long-tailed effects of focal loss \cite{lin2017focal}. Instant-Teaching \cite{zhou2021instant} uses a co-teaching~\cite{han2018co} framework to get accurate pseudo-labels. Combating noise~\cite{wang2021combating} alleviates the influences of noise pseudo-label by considering the uncertainty. MA-GCP~\cite{li2022semi} improves the consistency between the feature of the pseudo-labels and its corresponding global class feature. MUM~\cite{kim2022mum} utilizes the Interpolation-regularization (IR) and proposes a more effective strong-augmentation to improve the effectiveness of the pseudo-labels.   
In contrast, we consider a more challenging setting where the labeled data and unlabeled data come from different domains, causing semi-supervised object detection more difficult.

\noindent  \textbf{Cross-domain Object Detection.}
Cross-domain object detection \cite{chen2018domain,feng2020deep,saito2019strong,zhu2019adapting,cai2019exploring,hsu2020progressive,yu2019unsupervised} assumes the source data and target data have different data distributions and emphasizes the high performance on target data. However, they focus on the target domain performance and ignore the performance drop on the source domain, and are limited to the single target domain. Beyond that, multi-target domain adaptation aims to address domain shifts between the source and multiple target domains. Existing studies~\cite{yu2018multi,nguyen2021unsupervised,gholami2020unsupervised,isobe2021multi} focus on multiple domain shifts in classification tasks. 
However, directly applying existing multi-target domain adaptation methods to domain-inconsistent SSOD is unfavorable, since they ignore the long-tailed class imbalance within each domain~\cite{zhang2021deep,zhang2022test} and class distribution shifts among domains in SSOD. Moreover,  to the best of our knowledge, we are the first to explore the multi domains class distribution shifts problem in semi-supervised object detection.

\noindent \textbf{Discussion.} Overall, the domain-inconsistent SSOD task in this paper is different from cross-domain object detection and multi-target domain adaptation.
Specifically, domain-inconsistent SSOD considers multiple unlabeled domains and seeks to simultaneously handle co-variant shifts and class distribution shifts among these domains. In contrast, cross-domain object detection only considers a single target domain, while multi-target domain adaptation ignores the class distribution shifts. Hence,     domain-inconsistent SSOD is more challenging.

\begin{figure*}[ht]
\centering
\includegraphics[scale=0.65]{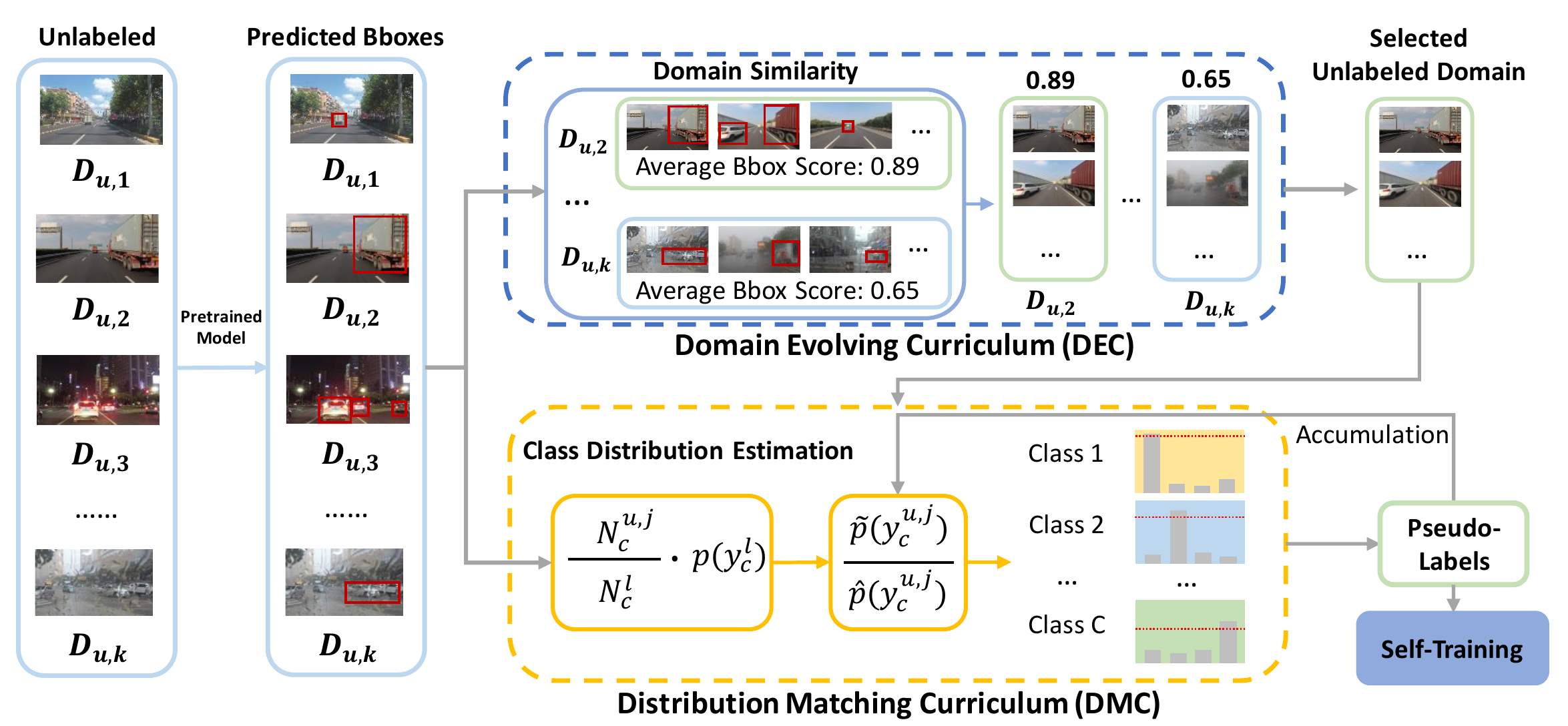}
\caption{Illustration of the proposed DucTeacher. DucTeacher proposes two curricula, DEC and DMC, to obtain accurate and unbiased pseudo-labels. DEC removes noise at the training data level and DMC removes the noise at the pseudo-label level. DMC adjusts the thresholds for each class by the ratio of accumulated pseudo-label distribution $\widetilde{p}(y^{u, j}_c)$ to the estimated class distribution of each domain $\hat{p}(y^{u, j}_c)$, which is estimated by the ratio of predicted bboxes numbers in each unlabeled domain $N_c^{u, j}$ to the numbers in the labeled domain $N_c^{l}$. Two curricula combine with each other and work for the same aim of providing accurate and unbiased pseudo-labels.}
\label{DucTeacher}
\end{figure*}

\section{Problem Definition}
\label{sec:setting}

As discussed in Sec.~\ref{sec:intro}, existing SSOD methods~\cite{jeong2019consistency, sohn2020simple, zhou2021instant, liu2021unbiased} are mainly limited in the IID setting, where the labeled and unlabeled data come from the same distribution.
Considering the practical situation in autonomous driving, we target a novel setting named \textbf{domain-inconsistent SSOD}, as shown in Fig.~\ref{Domain-Inconsistent}.
Specifically, there are domain shifts in the data, including (1) the data distribution shift, and (2) the class distribution shift between the labeled and unlabeled data.
Here, data distribution shift refers to the difference in pixel-level data distribution, such as daytime versus night.
On the other hand, the class distribution shift refers to the difference in the class distribution in each domain. 
As illustrated in Fig.~\ref{intuition}(b), the ``Daytime, Highway, Overcast'' domain has very few pedestrians while the ``Night, City street, Clear'' domain often has more pedestrians.
Meanwhile, we assume that only data from a few domains are annotated, while data from most other domains are unmarked. 
This is common in practical autonomous driving as the labeled data are limited and cannot cover many domains~\cite{han2021soda10m}.

Let $\mathcal{D}_L = \{{(x_i^{l}, y_i^{l})}\}_{i=1}^{N_l}$ represent the labeled set including $N_l$ labeled samples, where $y_i^{l} = \{{c_i^{l}, b_i^{l}\}}$ denotes the label that contains both the category label and the bounding box coordinate label. Let $\mathcal{D}_U = \{\mathcal{D}_{u, 1}, \mathcal{D}_{u, 2}, ..., \mathcal{D}_{u, k}\}$ be the unlabeled data of $k$ domains, which may include the labeled domain. For each domain $\mathcal{D}_{u, j}$, $j = {1, 2, ..., k}$, $\mathcal{D}_{u, j} = \{{x_i^{u, j}}, j\}_{i=1}^{N_{u, j}}$ represents $N_{u, j}$ unlabeled data in this domain.
There is no bounding box and class annotation for $\mathcal{D}_{u, j}$, but the domain index $j$ is assumed to be available. The domain index is easy to collect in practical applications, which represents the acquisition environment of unlabeled images (e.g., the location is \emph{City street}, the period is \emph{Daytime}, and the weather is \emph{Rainy}). 
Under the above domain-inconsistent SSOD setting,
the data distribution shift can be denoted as $p(x^{l}) \neq p(x^{u, j})$, that is, the distribution of the labeled data $x^{l}$ is different from that of the unlabeled data $x^{u, j}$.
On the other hand, the class distribution shift is denoted as $p(y^{l}) \neq p(y^{u, j})$, i.e., the class distributions of each domain are different.

\noindent \textbf{Challenges in domain-inconsistent SSOD.} The above domain-inconsistent setting introduces two challenges for SSOD.
First, the data distribution shifts would let the model trained on the labeled domain predict inaccurate pseudo-labels for the unlabeled domains with large distribution gaps. As shown in Fig.~\ref{intuition}(a), different levels of data distribution gaps result in different precision of the predicted pseudo-labels.
For instance, the model trained on the labeled images of daytime tends to make more mistakes for the unlabeled images at night, compared to the unlabeled images at dusk.  
Second, the class distribution shifts among domains make it difficult to obtain unbiased pseudo-labels for the unlabeled domains.
Specifically, the pseudo-labels would bias to the head classes in the labeled domain. A common practice to alleviate the biased pseudo-label distribution is to match the pseudo-label distribution to the class distribution of labeled data~\cite{berthelot2019remixmatch, he2021re}. However, when the unlabeled data come from multiple different domains with different class distributions, using the class distribution of the labeled data as a reference is inappropriate.

\section{Method}


In this section, we propose a dual-curriculum strategy named DucTeacher to eliminate noisy pseudo-labels in the data level and class level, as shown in Fig.~\ref{DucTeacher}.
First, we develop Domain Evolving Curriculum (DEC) in Sec.~\ref{sec:DEC} to introduce unlabeled data from different domains progressively according to the domain similarity, which aims to avoid noisy pseudo-labels on data from hard domains.
Second, we propose Distribution Matching Curriculum (DMC) in Sec.~\ref{sec:DMC}, a curriculum learning strategy to select pseudo-labels of different classes by matching the pseudo-label distribution and ground truth class distribution.

\subsection{Preliminary}

Existing state-of-the-art SSOD methods \cite{sohn2020simple, liu2021unbiased, zhou2021instant} usually adopt a Teacher-Student Mutual Learning framework. A student model is trained with combined loss function $\mathcal{L} = \mathcal{L}_s + \mathcal{L}_u$ with the supervised loss $\mathcal{L}_s$ and the unsupervised loss $\mathcal{L}_u$,
\begin{equation}
\begin{split}
  \mathcal{L}_s =  \frac{1}{N_{l}} \sum\nolimits_{i} \mathcal{L}_{cls}(x_i^l, y_i^l) + \mathcal{L}_{reg}(x_i^l, y_i^l),
\end{split}
\end{equation}
\begin{equation}
  \mathcal{L}_u = \frac{1}{N_{u}} \sum\nolimits_{i} \mathcal{L}_{cls}(x_i^u, y_i^u) + \mathcal{L}_{reg}(x_i^u, y_i^u),
\end{equation}
where $\mathcal{L}_{cls}$ and $\mathcal{L}_{reg}$ represent the classification loss and regression loss, respectively, $y_i^l$ is the annotation of the labeled image $x_i^l$, and $y_i^u$ is the pseudo-labels generated by teacher model on unlabeled data $x_i^u$.
Before the Teacher-Student Mutual Learning stage, the teacher model is pre-trained on the labeled set $\mathcal{D}_L$, while in the mutual learning stage, the teacher model is updated by exponential moving average (EMA) mechanism.

The quality of pseudo-labels $y_i^u$ is important for SSOD. Nevertheless, in domain-inconsistent SSOD, the input data distribution shifts and class distribution shifts would cause the teacher model to produce inaccurate and biased pseudo-labels $y_i^u$. To tackle this problem, we proposed two curricula, DEC and DMC, to provide reliable pseudo-labels $y_i^u$.

\subsection{Domain Evolving Curriculum}
\label{sec:DEC}
As shown in Fig.~\ref{intuition}(a), learning in domains with different domain similarities would produce pseudo-labels with noise at different levels, moreover, Fig.~\ref{intuition}(a) shows the teacher model produces unreliable pseudo-labels $y_i^u$ for domains with low domain similarity.
Based on the observation and the idea of learning easy samples first, DEC aims to learn similar domains first and learn the dissimilar domains after the model performance has been improved, which avoids the influence of noisy pseudo-labels produced in dissimilar domains. 

\noindent \textbf{Domain Similarity.}
We define the domain similarity to measure the difficulty of each unlabeled domain. Prior work \cite{cascante2020curriculum} shows the max prediction score can be regarded as a metric of uncertainty in image classification. In this work, we propose to use the average bboxes score $S_{u, j}$ for different unlabeled domains $D_{u, j}, j = 1, 2, 3, ..., k$ to measure the domain similarity. 
\begin{equation}
   S_{u, j} = \frac{1}{N_{u, j}}\sum_{i = 1}^{N_{u, j}} \bar{f}_{\theta}(y_{max}|x_{i}^{u, j}) ,
\end{equation}
where $N_{u, j}$ represents the image number in domain $\mathcal{D}_{u, j}$, $\bar{f}_{\theta}(y_{max}|x_{i}^{u, j})$ represents the average of max class probability of the predicted bboxes for each image from the unlabeled domain $x_{i}^{u, j}$. A domain with a high average bboxes score represents its high self-entropy and can be regarded as an easy domain for the model. 

\noindent \textbf{Domain Evolving Training.}
After computing the similarity for each unlabeled domain. To eliminate the influence of noisy pseudo-labels in dissimilar domains, DEC selects unlabeled data from domain $\mathcal{D}_{u, j}$ with the high domain similarity $S_{u, j}$ first then the dissimilar domains. 

\begin{table*}[t]
  \centering
  \normalsize
   \caption{ Comparison of mAP for different semi-supervised methods on SODA10M. The value in brackets represents the mAP improvement compared to the supervised model. }
   
  \resizebox{1.8\columnwidth}{!}{\begin{tabular}{cccccccccc}
    \toprule
    Method & mAP & $AP_{50}$ & $AP_{75}$ & Car & Truck & Pedestrian & Cyclist & Tram & Tricycle \\
    \midrule
    Supervised-only  &  37.9  & 61.6 & 40.4 & 58.3 & 43.2 & 31.0 & 43.2 & 41.3 & 10.5   \\
    \hline
    STAC \cite{sohn2020simple} &  42.8 \textcolor{red}{\footnotesize{(+ 4.9)}} & 64.8 & 46.0 & 63.4 & 47.5 & 35.7 & 46.4 & 44.4 & 19.6   \\
    UMT \cite{deng2021unbiased} &  44.7 \textcolor{red}{\footnotesize{(+ 6.8)}} & 67.5 & 48.2 & 65.1 & 49.9 & 34.6 & 48.1 & 50.2 & 14.3   \\
    MT-MTDA \cite{nguyen2021unsupervised} &  45.2 \textcolor{red}{\footnotesize{(+ 7.3)}} & 70.4 & 49.4 & 68.6 & 51.8 & 32.4 & 47.5 & 49.4 & 12.5   \\
    Unbiased Teacher \cite{liu2021unbiased} &  46.2 \textcolor{red}{\footnotesize{(+ 8.3)}}  & 70.1 & 50.2 & 67.9 & 53.9 & 33.8 & 50.2 & 55.2 & 16.4   \\
    MUM \cite{kim2022mum} &  45.9 \textcolor{red}{\footnotesize{(+ 8.0)}}  & 71.2 & 49.8 & 66.3 & 53.4 & 35.5 & 48.0 & 48.8 & 23.1   \\\rowcolor{Gray}
    DucTeacher w/o DMC    &  47.3 \textcolor{red}{\footnotesize{(+ 9.4)}} & 72.1 & 51.7 & 66.9 & 53.6 & 36.6 & 50.3 & 55.8 & 20.0   \\\rowcolor{Gray}
    DucTeacher w/o DEC    &  48.1 \textcolor{red}{\footnotesize{(+ 10.2)}} & 73.3 & 52.2 & 67.0 & 53.9 & 37.1 & 50.5 & 55.7 & 21.3   \\\rowcolor{Gray}
    DucTeacher (ours)   &  \textbf{48.4}  \textcolor{red}{\footnotesize{(+ 10.5)}} & 73.5 & 52.4 & 68.7 & 54.3 & 37.9 & 50.9 & 56.6 & 19.0   \\
    \bottomrule
  \end{tabular}}
  \label{tab:SODA10M}
\end{table*}

\subsection{Distribution Matching Curriculum}
\label{sec:DMC}
Current SSOD algorithms~\cite{liu2021unbiased,yang2021interactive,zhou2021instant,kim2022mum,li2022semi,wang2021combating} usually select pseudo-labels with confidence larger than a fixed pre-defined threshold. However, this strategy would cause the pseudo-labels $y_i^u$ bias to the head classes.
To reduce redundant head class pseudo-labels and encourage more neglected tail class pseudo-labels, DMC tries to match the ground truth class distribution and pseudo-label distribution by adjusting the class-specific and domain-specific thresholds at each iteration.

The intuition of DMC to dynamically adjust thresholds is that if the number of pseudo-labels produced by the teacher model is more than expected, the threshold would be raised and vice versa. The ratio of pseudo-label distribution and ground truth class distribution for each class $\widetilde{p}(y^{u, j}_c) / p(y^{u, j}_c)$ can be regarded as an indicator to raise or reduce the threshold for class $c$ in the domain $\mathcal{D}_{u, j}$. Moreover, because of the class distribution shifts in unlabeled domains, the proportion of each class in different domains is also different, which is shown in Fig.~\ref{intuition}(b) (e.g. Trams appearing on highway with a lower probability).  Hence, the threshold for the same class should also be different in different domains, which drives DMC to adjust thresholds $\textit{T}^{u, j}_c$ for each class at the domain level:
\begin{equation}
    \textit{T}^{u, j}_c = \tau + \mu \frac{\widetilde{p}(y^{u, j}_c)}{p(y^{u, j}_c)},
\end{equation}
where $\textit{T}^{u, j}_c$ represents the threshold for class $c$ in domain $\mathcal{D}_{u, j}$, $\tau$ is the pre-defined high threshold, and $\mu$ is the scale factor. $p(y^{u, j}_c)$ is the ground truth class distribution of class $c$, which represents the proportion of class label $c$ among all the ground truth labels in domain $\mathcal{D}_{u, j}$. The pseudo-label distribution $\widetilde{p}(y^{u, j}_c)$ is computed by accumulating the class number of model's predictions on the unlabeled data over the training course,
\begin{equation}
    \widetilde{p}(y^{u, j}_c) = \frac{1}{N^{u,j}_{p}} \sum_{i=1}^{N^{u,j}_{p}} \mathbbm{1} (f_{\theta_t}(y_c|x_i^{u, j}) > \textit{T}^{u, j}_c) \cdot \mathbbm{1}(c=C),
\end{equation}
\begin{equation}
    \textit{C} = argmax(f_{\theta_t}(y|x_i^{u ,j})),
\end{equation}
where $\widetilde{p}(y^{u, j}_c)$ is the cumulative pseudo-label distribution of class $c$ in domain $\mathcal{D}_{u, j}$ and is changed at each training iteration. $\widetilde{p}(y^{u, j}_c)$ can be regarded as the proportion of pseudo-labels of class $c$ over all the class in domain $D_{u,j}$. $f_{\theta}(y_c|x_i^{u, j})$ is model's prediction of the unlabeled data $x_i^{u, j}$ and $N^{u,j}_{p}$ is the number of pseudo-labels in domain $D_{u,j}$ over the training course.

\noindent \textbf{Estimating Class Distribution.}
However, the ground truth class distribution $p(y^{u, j}_c)$ in the unlabeled domain $\mathcal{D}_{u, j}$ is unavailable and because of the class distribution shifts between different domains, as shown in Fig.~\ref{intuition}(b), the class distribution of labeled data $p(y^{l}_c)$ is not an accurate estimation for that of unlabeled data. To resist the class distribution shifts, we propose a simple yet effective method to estimate the accurate class distribution for different unlabeled domains. 
First, by evaluating the unlabeled data with the model pre-trained on a labeled domain $\mathcal{D}_l$, we can obtain the predicted bboxes on the unlabeled data from all the domains $\mathcal{D}_{u, j}, j=1, 2, ..., k$. The accurate class distribution for each unlabeled domain $\mathcal{D}_{u, j}$ is computed as
\begin{equation}
    \hat{p}(y_c^{u, j}) = p(y_c^{l}) \cdot \frac{\mathcal{N}_c^{u, j}}{\mathcal{N}_c^{l}},
\label{con:estimation}
\end{equation}
where $\hat{p}(y_c^{u, j})$ is the estimated class distribution of unlabeled domain $\mathcal{D}_{u, j}$ for class $c$, $p(y_c^{l})$ is the ground truth class distribution of labeled data for class $c$, $\mathcal{N}_c^{l}$ and $\mathcal{N}_c^{u, j}$ are the number of predicted bboxes on the labeled domain $\mathcal{D}_{l}$ and the unlabeled domain $\mathcal{D}_{u, j}$ for class $c$. The proposed estimation method is based on this assumption, if the model is biased, the predicted results in each domain have the same bias. The proposed estimation method in Eqn.~\eqref{con:estimation} can remove the bias by division and after removing the bias of model's predictions, $\mathcal{N}_c^{u, j} / \mathcal{N}_c^{l}$ represents the scale ratio compared to $p(y_c^{l})$ .  Experiments in Sec.~\ref{sec:ablation} show the proposed estimation method can get much more accurate class distribution of the unlabeled data, and the estimated class distributions can help DMC adjust the threshold more accurately.


\begin{table*}[t]
    \fontsize{8}{8}\selectfont
    \caption{Comparison of mAP for different semi-supervised methods on SODA10M detailed in each domain. `-' means no validation image in this domain.}
    \centering
    \resizebox{1.8\columnwidth}{!}{
    \begin{tabular}{c|c|ccc|ccc|cc}
        \toprule
        \multirow{2}{*}{Model} & \multirow{2}{*}{Overall mAP} & \multicolumn{3}{c|}{City street (Car)} & \multicolumn{3}{c|}{Highway (Car)} & \multicolumn{2}{c}{Country road (Car)} \\
        \cmidrule{3-10}
         & & Clear & Overcast & Rainy & Clear & Overcast & Rainy & Clear & Overcast  \\
        \midrule
        \multicolumn{10}{c}{Daytime} \\
        \midrule
        Supervised & 43.1 & 70.0 & 64.9 & 56.6 & 68.3 & 65.9 & 65.9 & 69.4 & 63.5  \\
        \midrule
        STAC \cite{sohn2020simple} & 45.3$^{\textcolor{red}{+2.2}}$ & 74.2 & 69.6 & 58.0 & 71.7 & 70.3 & 70.7 & 75.2 & 69.8  \\
        UMT \cite{deng2021unbiased} & 45.1$^{\textcolor{red}{+2.0}}$ & 73.4 & 67.5 & 56.9 & 68.5 & 68.7 & 68.2 & 70.2 & 64.7  \\
        MT-MTDA \cite{nguyen2021unsupervised} & 47.1$^{\textcolor{red}{+4.0}}$ & 71.8 & 66.0 & 52.9 & 68.3 & 67.8 & 69.8 & 74.5 & 67.5  \\
        Unbiased Teacher \cite{liu2021unbiased} & 47.7$^{\textcolor{red}{+4.6}}$ & 73.0 & 68.1 & 55.3 & 69.1 & 62.0 & 71.3 & 72.6 & 70.0  \\\rowcolor{Gray}
        DucTeacher (ours) & \textbf{49.6}$^{\textcolor{red}{+6.5}}$ & 76.7 & 68.5 & 55.6 & 69.5 & 70.0 & 71.6 & 73.5 & 69.1  \\
        \midrule
        \multicolumn{10}{c}{Night} \\
        \midrule
        Supervised & 21.1 & 36.3 & 37.7 & - & 37.5 & 37.3 & 79.5 & 38.9 & 72.8  \\
        \midrule
        STAC \cite{sohn2020simple} & 28.2$^{\textcolor{red}{+7.1}}$ & 45.5 & 46.8 & - & 46.2 & 45.6 & 83.7 & 47.2 & 75.4  \\
        UMT \cite{deng2021unbiased} & 35.9$^{\textcolor{red}{+14.8}}$ & 58.4 & 59.7 & - & 58.7 & 60.2 & 81.1 & 60.4 & 72.2  \\
        MT-MTDA \cite{nguyen2021unsupervised} & 37.1$^{\textcolor{red}{+16.0}}$ & 60.4 & 61.2 & - & 60.7 & 62.2 & 80.6 & 62.4 & 73.6  \\
        Unbiased Teacher \cite{liu2021unbiased} & 39.7$^{\textcolor{red}{+18.6}}$ & 65.3 & 66.2 & - & 66.2 & 67.2 & 83.6 & 67.5 & 75.2  \\\rowcolor{Gray}
        DucTeacher (ours) & \textbf{40.7}$^{\textcolor{red}{+19.6}}$ & 65.3 & 67.0 & - & 66.8 & 67.4 & 84.3 & 67.7 & 76.5  \\
        \bottomrule
    \end{tabular}}
    \label{tab:domain_all}
\vspace{-0.05cm}
\end{table*}

\section{Experiments}
\subsection{Experimental Settings}
\paragraph{Datasets.} 
We first benchmark the proposed method on the recently proposed autonomous driving dataset SODA10M~\cite{han2021soda10m}. There are six class labels in SODA10M (i.e., Car, Truck, Pedestrian, Tram, Cyclist, Tricycle). It has 5,000 labeled data with 41,110 annotated bboxes from a single domain (i.e., Daytime, City street, Clear) and 10M unlabeled data from 48 domains \cite{han2021soda10m}. Following~\cite{han2021soda10m}, we use one million unlabeled data in SODA10M for SSOD. The validation set of the SODA10M has 5k data with annotated bboxes from 15 domains. Moreover, to verify the generalization of the proposed DucTeacher, we also conduct the experiments on MS-COCO \cite{lin2014microsoft} following \cite{liu2021unbiased}.

\noindent  \textbf{Baselines and evaluation metrics.} 
In this work, we adopt STAC \cite{sohn2020simple}, Unbiased Teacher \cite{liu2021unbiased}, CSD \cite{jeong2019consistency}, and Instant-Teaching \cite{zhou2021instant} as our baseline SSOD methods. 
We also consider the supervised baseline where the model is trained without unlabeled data. 
We use $AP_{50:95}$ (denoted as mAP) as the evaluation metric, which averages the ten AP values over $AP_{50}$ to $AP_{95}$.

\noindent \textbf{Implementation details.}
We implement our DucTeacher based on Detectron2 \cite{wu2019detectron2}. For a fair comparison, we follow STAC \cite{sohn2020simple} and Unbiased Teacher \cite{liu2021unbiased} to use Faster-RCNN with FPN \cite{lin2017feature} and ResNet-50 backbone \cite{he2016deep} as the object detector. We adopt the teacher-student mutual learning framework as Unbiased Teacher~\cite{liu2021unbiased}, which trains the student model with Focal loss ~\cite{lin2017focal} and updates the teacher model with EMA.
The EMA rate is $\alpha=0.9996$. The base pre-defined threshold in DucTeacher $\tau$ is set as 0.7, and the scale factor $\mu$ is set as 0.1. 
The learning rate is set as 0.01, and the max training iteration is set as 160k. For a fair comparison with existing works~\cite{sohn2020simple, zhou2021instant}, the batch size of training data is set as 16 for both the SODA10M and COCO datasets. The pre-trained model obtained in DucTeacher is trained on the labeled domain $\mathcal{D}_{l}$ with 20k iterations.
We conduct experiments with 8 Nvidia V100 GPU (32GB) cards with Intel Xeon Platinum 8168 CPU (2.70GHZ).
Details of the implementation and the hyper-parameters, can be found in Appendix~\ref{app:Implementation} and Appendix~\ref{app:HPO}, respectively. 

\begin{figure}[t]
\centering
\subcaptionbox{Unbiased Teacher. \label{subfig:error by Unbiased Teacher}}
{\includegraphics[width=1.5in]{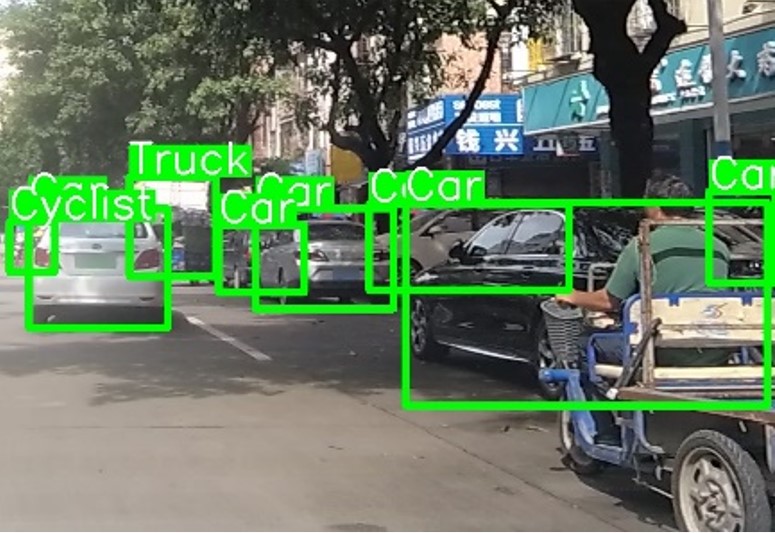}}
\hspace{1mm}
\subcaptionbox{DucTeacher (ours). \label{subfig:error by DucTeacher}}
{\includegraphics[width=1.5in]{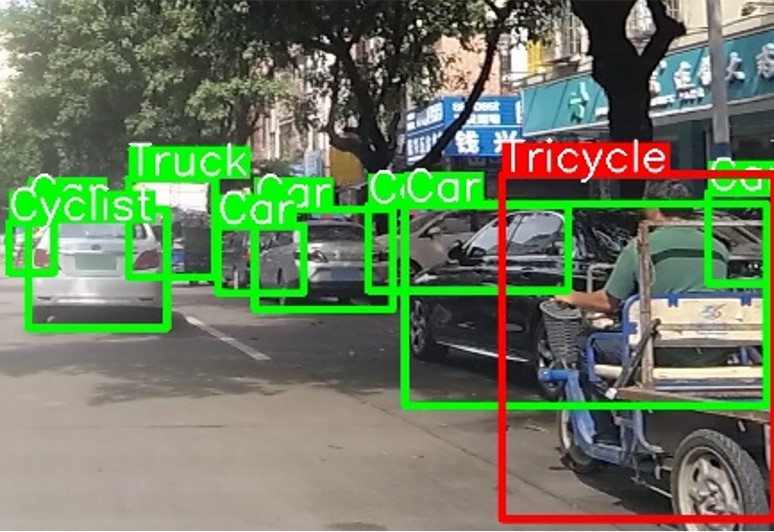}}
\caption{Visualizations of the predicted pseudo-labels from Unbiased Teacher and DucTeacher. The green bboxes represent detected objects that appeared in both the pseudo-labels of Unbiased Teacher and DucTeacher. The red bboxes represent the detected objects by DucTeacher but are neglected by Unbiased Teacher.
\label{fig_error}}
\vspace{-0.2cm}
\end{figure}

\emph{Adapting to the COCO.} Unlike the SODA10M dataset, there is no domain label in COCO \cite{lin2014microsoft}. To adapt the proposed DucTeacher for classical SSOD on COCO, we modify the proposed DEC. First, we also pre-train a model only with the labeled data until 2,000 iterations then use the pre-trained model to evaluate the unlabeled data and obtain the average bboxes score for each image. After that, we sort the unlabeled data in descending order according to the average bboxes score and divide the unlabeled data into different phases. Unlabeled data in each phase can be regarded as a similar difficulty degree. Also, we use the class distribution of labeled data as the ground truth class distribution for distribution matching.

\begin{figure}[t]
\centering
\includegraphics[scale=0.35]{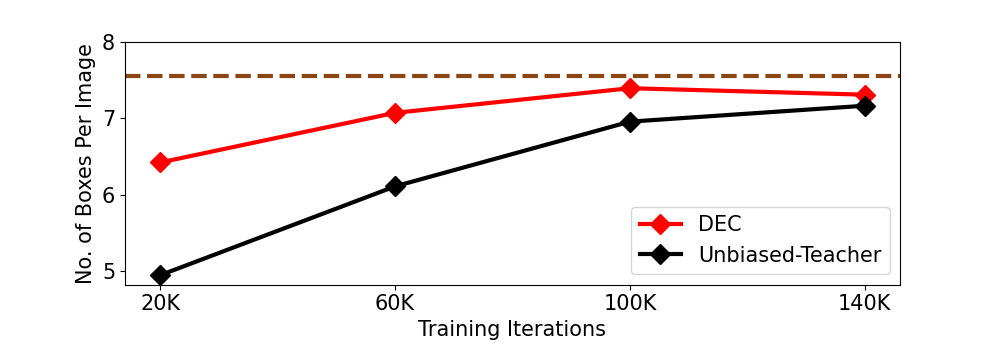}
\caption{Number of predicted boxes with scores larger than fix threshold 0.7 per image by Unbiased Teacher and DEC. The model trained with DEC produces more instances over all the training courses, especially in the early training stage. The brown dash line represents the average number of ground truth objects in the validation set.}
\label{DEC}
\vspace{-0.25cm}
\end{figure}

\subsection{Results}
\noindent \textbf{SODA10M.} We benchmark the proposed DucTeacher on the SODA10M dataset~\cite{han2021soda10m,wang2022memory}, compared with the supervised baseline and state-of-the-art SSOD methods ~\cite{sohn2020simple,liu2021unbiased}. Table \ref{tab:SODA10M} shows the superiority of DucTeacher, which improves 10.5mAP over the supervised baseline. Compared with the state-of-the-art SSOD method, Unbiased Teacher, DucTeacher can also improve the performance by about 2.2 mAP. Table~\ref{tab:SODA10M} also shows that compared to Unbiased Teacher, DucTeacher improves the AP performance for both the head class and the classes with poor performance. Specifically, DucTeacher outperforms Unbiased Teacher by 0.8 AP for head class ``Car'' and 4.1 AP, 2.6 AP for poor classes ``Pedestrian'' and ``Tricycle'', respectively. 

\noindent \textbf{COCO.}  We also evaluate the proposed DucTeacher in classical SSOD setting on COCO \cite{lin2014microsoft} following Unbiased Teacher \cite{liu2021unbiased}. Table \ref{tab:COCO} shows that DucTeacher achieves state-of-the-art performance under different ratios of labeled data. Note that 1$\%$ means that 1$\%$ of the total image are labeled, and the others are unlabeled. Although DucTeacher is focusing on tackling the noise prediction problem caused by multiple domains and class distribution shifts among this, DucTeacher can also improve detection in COCO.

\begin{figure}[t]
\centering
\subcaptionbox{Similar domain. \label{subfig:easy}}[.48\linewidth]
{\includegraphics[width=1.5in]{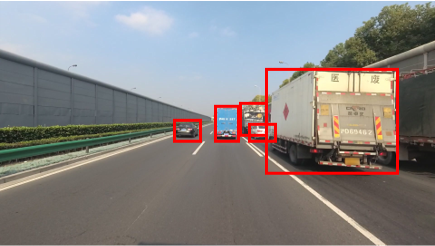}}
\subcaptionbox{Dissimilar domain. \label{subfig:hard}}[.48\linewidth]
{\includegraphics[width=1.5in]{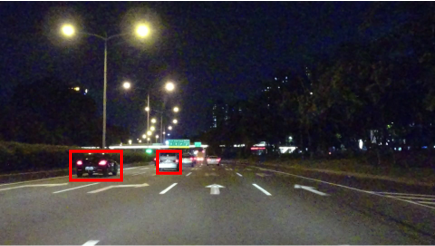}}
\caption{Visualizations of the pseudo-labels produced on similar and dissimilar domain. In the similar domain, there are fewer False Negatives errors (i.e., five ground truth objects versus five predicted bboxes). While in the dissimilar domain, objects are hard to detect and cause more False Negatives errors (five ground truth objects versus only two predicted bboxes). \label{DEC_mistake}} 
\vspace{-0.2cm}
\end{figure}

\subsection{Ablation Studies}
\label{sec:ablation}
\paragraph{Domain Evolving Curriculum.}
DEC is proposed to learn unlabeled data from a similar domain first, to avoid the noisy pseudo-label produced on the unlabeled data with a drastic data distribution shift. As shown in Fig.~\ref{intuition}(a), the domain similarity measured by the metric proposed in DEC is positively associated with the precision of pseudo-labels. Hence, we can use DEC to select the unlabeled data with a high domain similarity as easy unlabeled data to train first and avoid noisy pseudo-labels. Table~\ref{tab:SODA10M} shows that, with the help of DEC, the performance gain is 1.1 mAP compared to the Unbiased Teacher, showing the effectiveness of DEC. Moreover, Fig.~\ref{DEC} shows that with the help of DEC, the model would predict more instances per image, which represents DEC can make fewer False Negatives error. 
The False Negatives error seriously affects the self-training process since it would make the image lack annotations of some objects, which encourages the student model to predict background class for the unmarked objects and restrains the ground truth object class. 
Fig.~\ref{DEC_mistake} shows more False Negatives errors happen on dissimilar domain than the similar domain and when using only the detected objects as pseudo-labels, the student model would be encouraged to predict background classes for unmarked objects. Fig.~\ref{DEC} also shows the number of predicted boxes per image in DEC is much more than that in Unbiased Teacher in the early training iterations. This is because the proposed DEC removes the data from dissimilar domains in the early stage, which avoids making miss detection mistakes on data from dissimilar domains and then avoids the model learning unmarked object as a background class. 
\begin{figure}[h]
\centering
\includegraphics[scale=0.35]{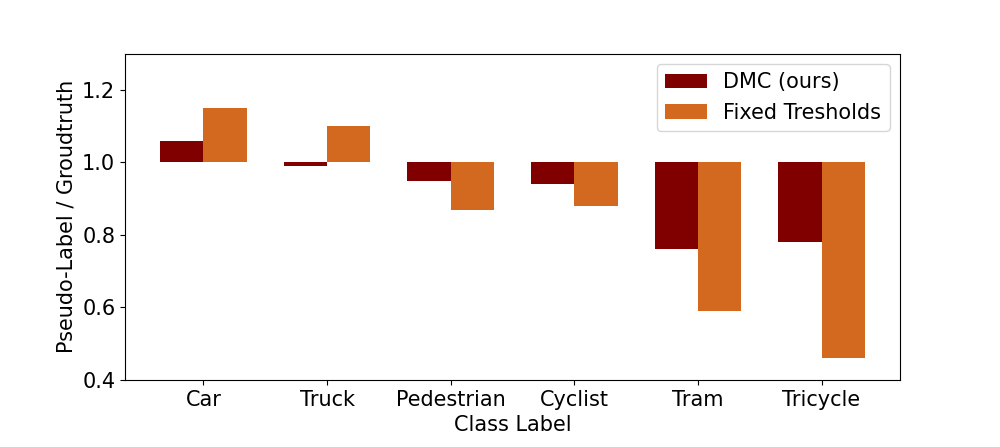}
\caption{The ratio of pseudo-label distribution and ground truth class distribution for each class. The ratio of pseudo-label distribution obtained by our method DMC and ground truth class distribution is much closer to 1, which means a less unbiased pseudo-label distribution. } 
\label{DMC match}
\vspace{-0.25cm}
\end{figure}
\begin{table}[t]
  \centering
  \large
  \caption{Comparison of mAP for different semi-supervised methods on MS-COCO under different labeled data ratios. The value in brackets represents the mAP improvement compared to the supervised model.}
  \resizebox{1\columnwidth}{!}{\begin{tabular}{cccccc}
    \toprule
    Method & 1$\%$ & 2$\%$ & 5$\%$ & 10$\%$ \\
    \midrule
    Supervised-only  & 9.05 & 12.70 & 18.47 & 23.86 \\
    \hline
    CSD \cite{jeong2019consistency}& 10.51 \textcolor{red}{\footnotesize{(+ 1.46)}} & 13.93 \textcolor{red}{\footnotesize{(+ 1.23)}} & 18.63 \textcolor{red}{\footnotesize{(+ 0.16)}} & 22.46 \textcolor{red}{\footnotesize{(- 1.40)}} \\
    STAC \cite{sohn2020simple} & 13.97 \textcolor{red}{\footnotesize{(+ 4.92)}} & 18.25 \textcolor{red}{\footnotesize{(+ 5.55)}} & 24.38 \textcolor{red}{\footnotesize{(+ 5.91)}} & 28.64 \textcolor{red}{\footnotesize{(+ 4.78)}} \\
    Instant-Teaching \cite{zhou2021instant} & 18.05 \textcolor{red}{\footnotesize{(+ 9.00)}} & 22.45 \textcolor{red}{\footnotesize{(+ 9.75)}}   & 26.75 \textcolor{red}{\footnotesize{(+ 8.28)}} &30.40 \textcolor{red}{\footnotesize{(+ 6.54)}}    \\
    Unbiased Teacher \cite{liu2021unbiased} & 19.60 \textcolor{red}{\footnotesize{(+ 10.55)}} & 23.64 \textcolor{red}{\footnotesize{(+ 10.94)}} & 27.85 \textcolor{red}{\footnotesize{(+ 9.38)}} & 30.90 \textcolor{red}{\footnotesize{(+ 7.04)}} \\\rowcolor{Gray}
    DucTeacher (ours)  & 20.35 \textcolor{red}{\footnotesize{(+ 11.30)}} & 24.18 \textcolor{red}{\footnotesize{(+ 11.48)}} & 28.23 \textcolor{red}{\footnotesize{(+ 9.76)}} & 31.21 \textcolor{red}{\footnotesize{(+ 7.35)}} \\
    \bottomrule
  \end{tabular}}
  \label{tab:COCO}
 \vspace{-0.22cm}
\end{table}

\noindent \textbf{Distribution Matching Curriculum.}
DMC is proposed to introduce pseudo-labels with the cut of dynamical thresholds for each class and obtain unbiased pseudo-labels. As shown in Fig.~\ref{DMC match}, using the fixed thresholds for each class to select pseudo-labels would produce redundant head class pseudo-labels and restrain the produce of tail class pseudo-labels, which causes a high ratio of pseudo-label distribution to ground truth class distribution for head class Car and low ratios for tail classes Tram and Tricycle. Compared with the Fixed thresholds, the ratio of pseudo-label distribution produced by DMC to ground truth class distribution is much closer to 1, which represents DMC can inhibit the production of over-confident pseudo-labels of head class, and encourage much more neglected pseudo-labels of tail class. Table \ref{tab:SODA10M} also shows with the effect of DMC, DucTeacher can achieve 1.4 AP improvement compared to Unbiased Teacher, moreover, improve 4.9 AP for tail class (Tricycle). Fig.~\ref{fig_error} shows DucTeacher can detect poor class objects neglected by Unbiased Teacher and this kind of predicted bboxes would further promote the learning course for poor classes in a self-training framework, which improves the recognition ability for poor classes.

\begin{table}[t]
  \centering
  \large
  \caption{Performance comparison for Baseline, DMC with class distribution of labeled data as reference and DMC with estimated class distribution as reference.}
  \resizebox{0.8\columnwidth}{!}{\begin{tabular}{llll}
    \toprule
    Method & mAP & AP50 & AP75 \\
    \midrule
    Baseline & 46.2 & 70.1 & 50.2 \\
    \hline
    DMC w/ Labeled  & 47.2 \textcolor{red}{\footnotesize{(+ 1.0)}} & 71.6 \textcolor{red}{\footnotesize{(+ 1.5)}} & 51.0 \textcolor{red}{\footnotesize{(+ 0.8)}}  \\
    DMC w/ Estimated  & \textbf{48.4} \textcolor{red}{\footnotesize{(+ 2.2)}} & \textbf{73.5} \textcolor{red}{\footnotesize{(+ 3.4)}} & \textbf{52.4} \textcolor{red}{\footnotesize{(+ 2.2)}}  \\
    \bottomrule
  \end{tabular}}
  \label{tab:classestimation}
\end{table}

\begin{figure}[h]
\centering
\includegraphics[scale=0.35]{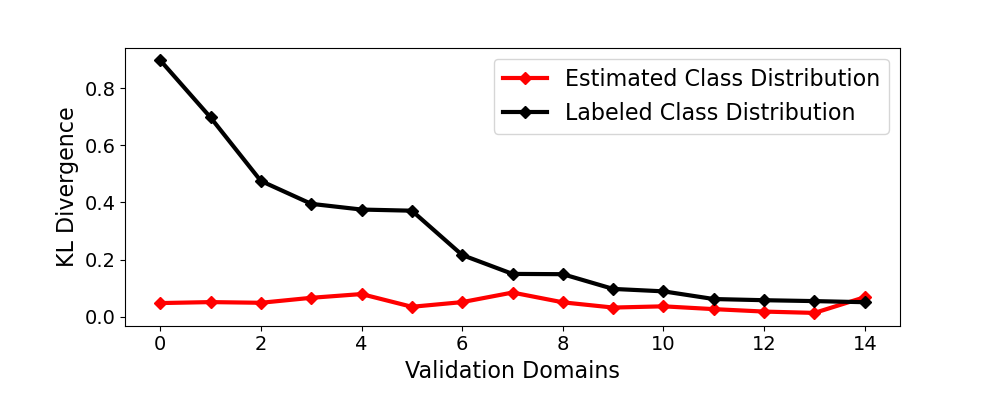}
\caption{KL divergence between the ground truth class distribution in the validation set and estimated class distribution (red line), class distribution of labeled data (black line). The Estimated Class Distribution has lower KL divergence than the Labeled Class Distribution, which means the estimated class distribution can be regarded as an accurate estimation for the ground truth class distribution of unlabeled domains.}
\label{KL divergence}
\vspace{-0.24cm}
\end{figure}

\noindent \textbf{Estimating  Class  Distribution.}
To rectify the pseudo-label distribution, we need to accurately estimate the class distributions of multiple unlabeled domains. To verify the effectiveness of the proposed class distribution estimation method, we estimate the class distribution of the validation set and observe the difference between estimated class distributions and ground truth class distributions. Fig.~\ref{KL divergence} shows the KL divergence between estimated class distributions and ground truth class distributions are lower than 0.2 for all the unlabeled domains. Moreover, Fig.~\ref{KL divergence} shows when existing class distribution shifts, class distribution of labeled data would have a high KL divergence with that of unlabeled domains (black line), which means regarding the class distributions of labeled data and unlabeled data as the same is inaccurate. Table \ref{tab:classestimation} shows the performance of calibrating the pseudo-label distribution by regarding the class distribution of labeled data as ground truth or using the estimated class distribution. Experiments show using DMC to calibrate the pseudo-label distribution by taking the class distribution of labeled data or the estimated class distribution as the reference can improve the performance by 1.0 mAP and 1.4 mAP, which shows using the estimated class distribution as reference is a more accurate choice to adjust thresholds for each class and improve performance.

\section{Conclusions}
In this paper, we introduce a practical semi-supervised object detection setting for autonomous driving, named as Domain-Inconsistent Semi-Supervised Object Detection, where the labeled data and unlabeled data come from multiple different domains. Moreover, we point out there are two serious problems, input data distribution shifts and class distribution shifts. Furthermore, we propose DucTeacher with two curricula, DEC and DMC, to obtain accurate and unbiased pseudo-labels. Experiments show our DucTeacher achieves satisfactory performance on both the domain-inconsistent SSOD dataset SODA10M and the classical SSOD dataset COCO. Beyond DucTeacher, how to design a more effective data curriculum strategy, such as without using domain labels, is interesting. Second, extending DucTeacher to the 3D autonomous driving scene is also significant. Third, due to the large training cost of training an autonomous driving model, combining the advantages of semi-supervised object detection and continual object detection to improve both the precision and training efficiency is considerable. 

\section*{Acknowledgement}

We gratefully acknowledge the support of MindSpore, CANN (Compute Architecture for Neural Networks) and Ascend AI Processor used for this research.
{\small
\bibliographystyle{ieee_fullname}
\bibliography{egbib}
}

\clearpage
\newpage
\appendix

\section*{Appendix}
We organize the Appendix as follows:
\begin{itemize}
    \item In Appendix~\ref{app:setting}, we add some additional explanation for the Domain-Inconsistent Object Detection.
  \item In Appendix~\ref{app:Framework}, we introduce the mutual learning framework used in our DucTeacher.
  \item In Appendix~\ref{app:Implementation}, we demonstrate the implementation details about DucTeacher. 
  \item In Appendix~\ref{app:HPO}, we analyze the influence of the hyper-parameter,  thresholds $\tau$ and scale factors $\mu$, introduced in the DucTeacher.  
    \item In Appendix~\ref{app:SODA10M}, we show more results about SODA10M.  
  \item In Appendix~\ref{app:EMA}, we explain how DucTeacher can show a good cross-domain generalization ability. 
\end{itemize}

\section{Domain-Inconsistent Object Detection Setting}
\label{app:setting}
As exhibited in the main paper, the targeted setting, Domain-Inconsistent Object Detection is different from the Classical Semi-Supervised Object Detection. Fig.~\ref{Domain-Inconsistent} shows the difference for detail. In the Classical Semi-Supervised Object Detection, Labeled data and Unlabeled data are from the same data distribution. However, in the Domain-Inconsistent Semi-Supervised Object Detection, Labeled Data and Unlabeled data can from different different distribution. This distribution shift would cause two detailed challenges hindering the learning of unlabeled data and this influence analysis has elaborated in the main paper.

\section{Mutual Learning Framework}
\label{app:Framework}

Existing state-of-the-art SSOD methods \cite{sohn2020simple,liu2021unbiased,zhou2021instant} usually adopt a Teacher-Student Mutual Learning framework. Similar to the knowledge distillation~\cite{hinton2015distilling,yu2022multi}, a student model is partially supervised by the teacher model and is trained with combined loss function $\mathcal{L} = \mathcal{L}_s + \mathcal{L}_u$ with the supervised loss $\mathcal{L}_s$ and the unsupervised loss $\mathcal{L}_u$,
\begin{equation}
\begin{split}
  \mathcal{L}_s =  \frac{1}{N_{l}} \sum\nolimits_{i} \mathcal{L}_{cls}(x_i^l, y_i^l) + \mathcal{L}_{reg}(x_i^l, y_i^l),
\end{split}
\end{equation}
\begin{equation}
  \mathcal{L}_u = \frac{1}{N_{u}} \sum\nolimits_{i} \mathcal{L}_{cls}(x_i^u, y_i^u) + \mathcal{L}_{reg}(x_i^u, y_i^u),
\end{equation}
where $\mathcal{L}_{cls}$ and $\mathcal{L}_{reg}$ represent the classification loss and regression loss, respectively, $y_i^l$ is the annotation of the labeled image $x_i^l$, and $y_i^u$ is the pseudo-labels generated by teacher model on unlabeled data $x_i^u$.
Before the Teacher-Student Mutual Learning stage, the teacher model is pre-trained on the labeled set $\mathcal{D}_L$, while in the mutual learning stage, the teacher model is updated by exponential moving average (EMA) mechanism.

The quality of pseudo-labels $y_i^u$ is important for SSOD. Nevertheless, in domain-inconsistent SSOD, the input data distribution shifts and class distribution shifts would cause the teacher model to produce inaccurate and biased pseudo-labels $y_i^u$. To tackle this problem, we proposed two curricula, DEC and DMC, to provide reliable pseudo-labels $y_i^u$.

DucTeacher is based on the teacher-student mutual learning framework and the consistency pseudo-labeling strategy, which are also included in existing state-of-the-art SSOD methods. We introduce the technical details as follows. 

\noindent \textbf{Teacher-Student Mutual Learning.} In the mutual learning stage, the student model is trained with supervision of the ground truths and the pseudo-labels. The student model is updated by gradient descent, while the teacher model is updated by exponential moving average (EMA) mechanism,
\begin{equation}
    \theta_s \gets \theta_s + \frac{\partial \mathcal{L}}{\partial \theta_s},
\end{equation}
\begin{equation}
    \theta_t \gets \alpha\theta_t + (1-\alpha)\theta_s,
\end{equation}
where $\theta_t$, $\theta_s$ represent the parameter of teacher model and student model.

\noindent \textbf{Consistency Pseudo-Labeling.} Consistency Pseudo-Labeling produces pseudo-label based on both the consistency regularization and pseudo-labeling. It produces a pseudo-label on a weakly-augmented unlabeled image and screens out the pseudo-label with the high confidence score, which would be used as a target for the model fed with a strongly-augmented version of the same image. The detailed data augmentations used in this work are shown in Table \ref{tab:aug}.

\begin{table}[h]
  \centering
  \normalsize
  \caption{Used augmentations in DucTeacher.}
  \resizebox{0.99\columnwidth}{!}{\begin{tabular}{ccc}
    \toprule
    \multicolumn{3}{c}{\textbf{Weak Augmentation}}  \\
    \hline
    Process & Probability  &  Parameters \\
    \hline
    Horizontal Flip & 0.5 & - \\
    \hline
    \hline
    \multicolumn{3}{c}{\textbf{Strong Augmentation}}  \\
    \hline
    Process & Probability  &  Parameters \\
    \hline
    Grayscale       & 0.2 & - \\
    \hline
    GaussianBlur    & 0.5 & (sigma x, sigma y) = (0.1, 2.0) \\
    \hline
    CutoutPattern1  & 0.7 & scale=(0.05, 0.2), ratio=(0.3, 3.3) \\
    \hline
    CutoutPattern2  & 0.5 & scale=(0.02, 0.2), ratio=(0.1, 6) \\
    \hline
    CutoutPattern3  & 0.3 & scale=(0.02, 0.2), ratio=(0.05, 8) \\
    \bottomrule
  \end{tabular}}
  \label{tab:aug}
 \vspace{-0.30cm}
\end{table}

\section{Implementation details}
\label{app:Implementation}
We implement our DucTeacher based on Detectron2 \cite{wu2019detectron2}. For a fair comparison, we follow STAC \cite{sohn2020simple} and Unbiased Teacher \cite{liu2021unbiased} to use Faster-RCNN with FPN \cite{lin2017feature} and ResNet-50 backbone \cite{he2016deep} as the object detector. We adopt the teacher-student mutual learning framework as Unbiased Teacher~\cite{liu2021unbiased}, which trains the student model with Focal loss ~\cite{lin2017focal} and updates the teacher model with EMA.
The EMA rate $\alpha$ is 0.9996. The base pre-defined threshold in DucTeacher $\tau$ is set as 0.7, and the scale factor $\mu$ is set as 0.1. 
The learning rate is set as 0.01, and the max training iteration is set as 160k. The batch size of training data is set as 32 (16 for both labeled and unlabeled) for both the SODA10M and COCO datasets. The pre-trained model obtained in DucTeacher is trained on the labeled domain $\mathcal{D}_{l}$ with 2k iterations.
We conduct experiments with 8 Nvidia V100 GPU (32GB) cards with Intel Xeon Platinum 8168 CPU (2.70GHZ).

\noindent \textbf{Pre-train Stage.} For SODA10M, the Pre-train stage is using the labeled domain $D_l$ to get an initial model. For COCO, following the Unbiased Teacher \cite{liu2021unbiased} , we use a small amount of labeled data to pre-train a detector with 2000 iterations for fair comparisons. 

\noindent \textbf{DEC for SODA10M.} For SODA10M, according to the similarity score provided by DEC, we divide the whole unlabeled training set into 4 subsets. The earlier trained subset has a higher similarity score than that of subset trained later. 

\noindent \textbf{Adapting to the COCO.} Unlike the SODA10M dataset, there is no domain label in COCO \cite{lin2014microsoft}. To adapt the DucTeacher for classical SSOD on COCO, we modify the proposed DEC. First, we also pre-train a model only with the labeled data until 2,000 iterations then use the pre-trained model to evaluate the unlabeled data and obtain the average bboxes score for each image. After that, we sort the unlabeled data in descending order according to the average bboxes score and divide the unlabeled data into different phases. Unlabeled data in each phase can be regarded as a similar difficulty degree. Also, we use the class distribution of labeled data as the target for distribution matching.

\section{Hyper-parameters}
\label{app:HPO}

In this appendix, we study the hyper-parameters introduced by the DucTeacher, including the threshold $\tau$ and the scale factor $\mu$. We report the results in Table \ref{tab:ablation1}, which gives the following observations. When the threshold $\tau$ and scale factor $\mu$ are set as 0.7 and 0.1, respectively, DucTeacher achieves the best performance (48.4mAP). For the threshold $\tau$, too low or too high threshold $\tau$ would both damage the performance, as a low threshold would cause noisy pseudo-labels and a high threshold would discard much useful information. For the scale factor $\mu$, too small scale factor $\mu$ would cause the dynamic thresholding strategy in DucTeacher hard to exert its effectiveness since too small scale factor $\mu$ would cause the thresholds for different categories in different domains almost the same, which cannot dynamically filter biased pseudo-labels. Too large scale factor $\mu$ would also cause an unstable dynamic threshold then affect performance. Since DucTeacher accumulates the pseudo-label distribution according to each training iteration, too large scale factor $\mu$ would cause DucTeacher dependent on the sampling of each batch too much then affect the performance.

\begin{table}[h]
\centering
 \caption{Ablation study on the effects of different pre-defined thresholds $\tau$ and different pre-defined scale factors $\mu$.}
 \resizebox{0.45\textwidth}{!}{\begin{tabular}{cccc||cccc}
  \toprule
  $\tau$ & mAP & $AP_{50}$ & $AP_{75}$ & $\mu$ & mAP & $AP_{50}$ & $AP_{75}$  \\
  \hline
  0.6 & 45.9 & 70.6 &50.0 & 0.05 & 47.1 & 71.8 &51.2\\
  \hline
  0.7 & 48.4 & 73.5 &52.4 & 0.10 & 48.4 & 73.5 &52.4 \\
  \hline
  0.8 & 44.7 & 69.3 &48.5 & 0.15 & 46.9 & 71.4 &50.9 \\
  \bottomrule
 \end{tabular}}
 \label{tab:ablation1}
\end{table}

 \begin{table}[h]
  \centering
  \caption{The cross-domain generalization ability of weak-stong augmentation and EMA mechanism. The table shows the ablation studies about mAP performance on weak-stong augmentation (DucTeacher w/o aug) and EMA mechanism (DucTeacher w/o EMA).}
  \resizebox{0.45\textwidth}{!}{\begin{tabular}{cccc}
    \toprule
    Method & Overall & Daytime & Night \\
    \midrule
    Supervised-only & 37.9 & 43.1 & 21.1 \\
    \hline
    UMT~\cite{deng2021unbiased} & 44.7 & 45.1 & 35.9 \\
    MT-MTDA~\cite{nguyen2021unsupervised} & 45.2 & 47.1 & 37.1 \\
    DucTeacher  & 48.4 & 49.6 & 40.7 \\
    \hline
    DucTeacher w/o aug & 39.1  & 44.9 & 22.4  \\
    DucTeacher w/o EMA & 35.7  & 42.1 & 16.3  \\
    \bottomrule
  \end{tabular}}
  \label{tab:weak-stong}
\end{table}

\section{More experimental results on SODA10M}
\label{app:SODA10M}
Table~\ref{tab:domain_all} further shows that the proposed DucTeacher can improve the mAP performance for almost all the domains. Moreover, compared with the state-of-the-art cross-domain object detection method UMT~\cite{deng2021unbiased} and our implemented multi-target domain adaptation method MT-MTDA~\cite{nguyen2021unsupervised}, the proposed method DucTeacher shows prominent superiority, which outperforms the UMT and MT-MTDA about 3.7mAP and 3.2mAP respectively, as shown in the main paper. Also, as shown in Table~\ref{tab:domain_all}, it is interesting that our DucTeacher performs better than UMT and MT-MTDA on both the daytime domains (source) and the night domains (target). The key to the great performance of our DucTeacher on the multiple unlabeled domains is the "weak-strong data augmentation" and "EMA mechanism", where the ablation experiments are shown in appendix~\ref{app:EMA}.

\noindent \textbf{How dose domain shifts affect semi-supervised object detection ?} In our main paper, the first thing is that models trained in a similar (easy) domain would perform poor in the dissimilar (hard) domain, which would cause the the produced pseudo-label contain lots of noise. Motivated by this, we design our DucTeacher which includes two curricula to handle the noise pseudo-label in two levels, the training order of data from different domains and the adjustment of pseudo-label thresholds for different domains. Except for these two perspectives, there is also an interesting distinctive problem in semi-supervised object detection, where models produce high-confidence pseudo-box in different levels for easy domains and hard domains. As shown in Fig.~\ref{DEC_mistake}, the high-confidence pseudo-box predicted in the similar (easy) domain is more than that in the dissimilar (hard) domain. The less predicted boxes would cause the False Negatives error in object detection, which causes that the model suppress the activation and tends to predict all the object as background. 


\section{Why dose DucTeacher have a good cross-domain generalization ability?}
\label{app:EMA}

In the main paper, we have shown that DucTeacher has a  good cross-domain generalization ability. In this appendix, we further analyze the reason behind it. In Table~\ref{tab:weak-stong}, we analyze the effectiveness of weak-strong augmentation and EMA mechanism for the cross-domain generalization ability of our DucTeacher. Compared with the supervised method, the main improvement of DucTeacher is in the Night domain, which is 19.6mAP improvement compared with the 6.5mAP improvement in the Daytime domain, as shown in Table~\ref{tab:weak-stong}. Moreover, DucTeacher also shows a higher cross-domain generalization performance compared with the existing state-of-the-art cross-domain object detection method UMT~\cite{deng2021unbiased} and multi-target domain adaptation method MT-MTDA~\cite{nguyen2021unsupervised}. We find the reason for the high cross-domain generalization ability of our DucTeacher is the combination of weak-strong augmentation and exponential moving average (EMA) mechanism. As shown in Table~\ref{tab:weak-stong}, without weak-strong augmentation or EMA mechanism, the performance drops greatly, especially in the Night domain, where the decreasing performance is 18.3mAP and 24.4mAP. With the combination of weak-strong augmentation and EMA mechanism, DucTeacher produces the pseudo-labels on the weakly-augmented images and the pseudo-labels would be used for the strongly-augmented images. This weak-strong consistency regularization guarantees DucTeacher's high cross-domain generalization ability. Besides, the EMA mechanism gradually updates the DucTeacher model to construct a temporal ensembles model of student model in different steps, which can produce more reliable pseudo-labels for self-training.

\end{document}